\documentclass[runningheads]{llncs}
\usepackage[T1]{fontenc}

\usepackage{graphicx}

\usepackage{multirow}
\usepackage{array}
\usepackage{graphicx}
\usepackage[table]{xcolor}
\usepackage{hyperref}
\usepackage{float}
\usepackage{subcaption}
\usepackage[many]{tcolorbox}
\usepackage{xcolor}
\usepackage{tikz}
\usepackage[linesnumbered,ruled,lined,boxed,vlined]{algorithm2e}

\usepackage{derivative}
\usepackage{multirow}

\usepackage{caption}
\definecolor{boxgrey}{HTML}{F3F3F3}

\newcommand{\hlbox}[2]{%
  \begin{center}%
    \fcolorbox{white}{boxgrey}{%
      \parbox{0.9\columnwidth}{\noindent \textbf{#1}. \textit{#2}}
    }%
  \end{center}%
}

\usepackage{fancyhdr}
\renewcommand{\headrulewidth}{0pt}
\begin{document}
\title{Effective Early Stopping of Point Cloud Neural Networks}

\author{Thanasis Zoumpekas\inst{1}\orcidID{0000-0002-3736-1155
} \and  Maria Salamó\inst{1,2}\orcidID{0000-0003-1939-8963} \and Anna Puig\inst{1,3}\orcidID{0000-0002-2184-2800}
}

\authorrunning{T. Zoumpekas et al.} 

\institute{WAI Research Group, Department of Mathematics and Computer Science, University of Barcelona, Barcelona, Spain \\
\email{\{thanasis.zoumpekas, maria.salamo, annapuig\}@ub.edu}
\\ 
\and
UBICS Institute, University of Barcelona, Barcelona, Spain \\
\and
IMUB  Institute, University of Barcelona, Barcelona, Spain}

\maketitle

\thispagestyle{plain}
\fancypagestyle{plain}{%
\fancyhf{}
\lfoot{\tiny{This preprint has not undergone peer review or any post-submission improvements or corrections. The Version of Record of this contribution is published in MDAI 2022, LNCS, vol 13408, Springer, Cham. Available online at \href{https://doi.org/10.1007/978-3-031-13448-7\_13}{https://doi.org/10.1007/978-3-031-13448-7\_13}.}}
\renewcommand{\headrulewidth}{0pt}%
}

\begin{abstract}

Early stopping techniques can be utilized to decrease the time cost, however currently the ultimate goal of early stopping techniques is closely related to the accuracy upgrade or the ability of the neural network to generalize better on unseen data without being large or complex in structure and not directly with its efficiency. Time efficiency is a critical factor in neural networks, especially when dealing with the segmentation of 3D point cloud data, not only because a neural network itself is computationally expensive, but also because point clouds are large and noisy data, making learning processes even more costly. In this paper, we propose a new early stopping technique based on fundamental mathematics aiming to upgrade the trade-off between the learning efficiency and accuracy of neural networks dealing with 3D point clouds. Our results show that by employing our early stopping technique in four distinct and highly utilized neural networks in segmenting 3D point clouds, the training time efficiency of the models is greatly improved, with efficiency gain values reaching up to 94\%, while the models achieving in just a few epochs approximately similar segmentation accuracy metric values like the ones that are obtained in the training of the neural networks in 200 epochs. Also, our proposal outperforms four conventional early stopping approaches in segmentation accuracy, implying a promising innovative early stopping technique in point cloud segmentation.


\end{abstract}

\section{Introduction}
\label{intro}

Since the popularity and the demand in 3D point cloud data analysis is constantly increasing, the rigorous evaluation of intelligent techniques dealing with such data is becoming a need. In particular, the appearance of 3D sensors, such as LIDAR and RGB-D cameras, among others, has favoured the creation of 3D-representations of areas (e.g., map) or objects (e.g., car). The set of data points in the $x$, $y$ and $z$ coordinated 3D space appearing in these 3D-representations is a point cloud that represents a 3D shape or object. Their speciality is derived by their noisy and irregular nature and because of this, analysis tasks, such as segmentation, that are common to deal efficiently and effectively in the 2D data domain, portray additional challenges in terms of efficiency and robustness in 3D.

Neural Networks (NN) are the most suitable machine learning algorithms to segment point cloud data due to their ability to handle and take advantage of the huge amount of points (i.e. millions in most cases) that a 3D point cloud dataset contains \cite{Bello2020Review:Clouds}. Different NN architectures have been proposed recently to segment inner structures of point cloud \cite{Qi2017PointNet:Segmentation,Qi2017PointNet++:Space,Yan2020PointASNL:Sampling}. However, recent studies show that training and evaluating these models are a greatly time-consuming task and, in some cases, the number of epochs and time spend is not proportional to the accuracy achieved \cite{Zoumpekas2021BenchmarkingSegmentation,Zoumpekas2022CLOSED:Learning}. Indeed, in this process, it is important not just to solely upgrade the accuracy-related metrics of the learning, such as accuracy, precision, recall or F1-score, but also to achieve a balance between efficiency and accuracy \cite{Zoumpekas2021BenchmarkingSegmentation}.

In this paper, we concentrate on 3D point cloud part segmentation analysis and on the upgrade of the trade-off between segmentation accuracy and efficiency of the learning process of NN models. By employing fundamental mathematical methodologies, we propose an algorithmic way that defines an early stopping criterion on the learning process of the models which aims to finish the learning process at an early point but maintaining an accurate enough model for making predictions on the test data. Our results show that by employing our early stopping technique to four of the most well-known neural networks in point cloud segmentation, the trade-off between training time efficiency and segmentation accuracy of the models is greatly improved. The models achieve in just a few epochs comparable segmentation accuracy metric values to the ones obtained in the training of the NN. Besides, the comparison with four conventional early stopping techniques in terms of obtained accuracy and loss analysis indicates that our proposal is a promising novel technique of early stopping in the NN models dealing with point cloud segmentation.

\section{Related work}
\label{sec:related_work}

The study of the ways to enhance the learning process of NN models to achieve higher accuracy and efficiency are major open issues. In the point cloud segmentation field, there are numerous studies dealing with advancements in the architectures or the learning parameters of the developed models to achieve higher segmentation accuracy \cite{Bello2020Review:Clouds,Zhang2019ACloud,Guo2020DeepSurvey,Liu2019DeepSurvey}. However, research must go beyond pure accuracy-metrics and another open issue of utmost importance is the efficiency of such deep learning models dealing with 3D point clouds, although it is still in its early steps of research. Recent studies highlight that the efficiency of 3D point cloud segmentation models is a serious concern for the community \cite{Zhang2019ACloud,Zoumpekas2021BenchmarkingSegmentation,Garcia-Garcia2018ASegmentation}. However, the majority of new and advanced deep learning models emphasize on the improvement of segmentation accuracy, providing almost no information on the models' efficiency \cite{Hegde2021PIG-Net:Segmentation,Liu2019Relation-shapeAnalysis,Qi2017PointNet++:Space,Thomas2019KPConv:Clouds}.

On the other hand, certain techniques of early stopping of the neural networks are utilized to handle either specific learning issues or time efficiency in the training phase. Caruana et al. \cite{Caruana2000OverfittingStopping} showed that when huge neural networks have learnt models that are similar to those learned by smaller ones, early stopping can be employed to stop their training process without significant loss in generalization performance.

Specifically, early stopping aims to stop the optimization of the training process of a neural network at an early stage in order to, firstly, mitigate the performance issues caused by overfitting, such as loss of generalization of the network, and secondly to improve the training time cost, i.e. time efficiency \cite{Prechelt2012EarlyWhen}. While there are many strategies for dealing with overfitting issues, such as regularization, network-reduction strategy, or data expansion, the early stopping techniques are the most used ones \cite{Ying2019AnSolutions}. Data science researchers mostly utilize early stopping techniques based on a threshold monitoring a loss function's values \cite{Prechelt2012EarlyWhen,Bai2021UnderstandingLabels}. For instance, the stopping of training when the error on the validation data is higher than the one recorded in the previous epoch or epochs. Although, the aforementioned technique is theoretically correct, there is always more than one local minimum in real validation error curves and, thus several stopping criteria utilize windows (or fixed intervals of epochs) capturing the evolution of validation error are employed in order to deal with this \cite{Prechelt2012EarlyWhen}. Recently, Bai et al. \cite{Bai2021UnderstandingLabels} proposed a progressive early stopping technique, in which they split a neural network into multiple parts and train them individually. Rather than employing traditional early stopping techniques, which require training the entire neural network at once, they train and optimize parts of a neural network by using early stopping criteria in those parts.

In summary, the majority of the related works propose early stopping techniques aiming to deal with learning issues of the models, such as overfitting, in order to lead to higher accuracy values in unseen data (test data), i.e. improve the model's generalization performance. They either rely on the whole network learning process (traditional early stopping) or in specific areas and segments of the NN (progressive early stopping). However, at the time of writing this paper, there are no studies dealing with early stopping of the models with a goal to provide a more efficient, in terms of time, learning process but also a highly accurate model. Thus, in an attempt to fill this gap we propose an effective early stopping aiming to get a model in a state that is highly accurate but also having spent a low amount of time in its training (efficient in run-time). For this, our approach focuses on the segmentation accuracy-related performance metrics instead of the loss function values (see Section \ref{sec:proposal}). We categorize our study in the traditional early stopping techniques, mainly because we do not split the architecture of the utilized neural networks into smaller parts.

\section{Early stopping of point cloud neural networks}
\label{sec:proposal}

This section presents, first, our early stopping criteria, and then, the algorithmic way to select the early stopping of the learning process of the models.

\subsection{Early stopping criteria}
\label{subsec:method}

We propose an automatic and online way of stopping the learning process of a point cloud part segmentation task based on the analysis of a stop-window containing the values of a monitored performance metric. Please, note that the monitored performance metric can be any segmentation accuracy-related metric. However, we have selected the $ImIoU$, i.e. the mean value of $IoU$ across all point cloud Instances, explained in \cite{Liu2019Relation-shapeAnalysis}, because it provides a general segmentation accuracy evaluation across all the available point cloud instances.

Initially, performance metric values $x$ are obtained by testing the NN in the whole test set after each training epoch. Indeed, in every epoch we train, validate and test the NN. The testing time after each epoch consists of an almost negligible cost compared to the training-validation phases, because it is a forward pass of the data to the model and, as Zhang et al.\cite{Zhang2019ACloud} showed, it is on the scale of milliseconds or even seconds. 

Assuming that the performance metric values are samples that can be approximated by a continuous smooth and differentiable (monotonic) function $f(x)$ that converges to a maximum upper bound value, we can calculate the first and second derivative values of $f(x)$, being $f'(x)$ and $f''(x)$ respectively. We approximate the first and second derivatives of $f(x)$ using the formulation appearing in Equations \ref{eq:f'} and \ref{eq:f''} respectively.

\begin{equation}
\label{eq:f'}
    f'(x) = \frac{f(x+h) - f(x)}{h} \quad ,
\end{equation}

\begin{equation}
\label{eq:f''}
    f''(x) = \frac{f'(x+h) - f'(x)}{h} \quad ,
\end{equation}

\noindent where $h$ is the distance between the data points. The data points are evenly spaced, because we have one performance metric value at every epoch ($h = 1$).  

First, we define a \textbf{window}, $w_{ij}$, to be the set of sampled values of $f(x)$ in the interval $(x_i, x_j)$, with $x_i < x_j$, and considering that $f(x_i)$, $f(x_j)$ are local maximum and minimum respectively. Please, note that $x_i$ refers to the $i$-th and $x_j$ refers to the $j$-th epoch. Moreover we define $size(w_{ij}) = j-i$, as the length of the interval domain, and $range(w_{ij}) = \{f(x_i) - f(x_{i+1}),..., f(x_{j-1}) - f(x_j)\}$ the interval range vector. The $range(w_{ij})$ is defined as a vector containing all the differences between each pair of consecutive performance metric values inside the window.

The underlying function of an accuracy-related metric oscillates between local minimum and maximum values. Note that we aim to define intervals to detect when the function oscillates less and converges to a certain value. To do this we could use: (i) Fixed-size intervals (i.e. windows of fixed-size) but they are not adaptive, i.e. taking into account the oscillations of the function, (ii) Adaptive intervals, where we detect the oscillations between local minimum and maximum values to observe the amplitude of the window. Indeed, it could be between a local maximum and a local minimum. In our case, we opted for the latter. The rationale behind our decision is that by starting the stop window at a local maximum, the model reached a peak in its accuracy, thus it seems like a spot to initiate an analysis. Then, we stop at the local minimum after the local maximum, because the model reached a trough spot or alternatively a negative peak, implying that no higher accuracy can be obtained within this interval.

Fundamentally, the \textbf{local minimum} of a function can be found where the first derivative of the function is equal to zero, i.e. $f'(x) = 0$, and the second derivative in this exact spot is greater than zero, i.e. $f''(x) > 0$. The approximation of \textbf{local maximum} is likewise the same ($f'(x)=0$) but in this case the second derivative of the function in that spot should be less than zero ($f''(x) < 0$).

Therefore, we define the \textbf{stop-window} as the window, $w_{ij}$, that fulfils certain conditions. In our case, we utilize the following conditions:
\begin{enumerate}
    \item $size(w_{ij}) \geq N$, with $N$ taking values within $[2, maxEpochs-1]$, and
    \item $\forall k$  $\in range(w_{ij}): $ $ |k| < D$, with $D$ taking values in the range of $(0, 2]$.
\end{enumerate}

The $size(w_{ij})$ is defined as the minimum distance in epochs between the local maximum and local minimum point. For instance, if $N = 4$, then the distance between the local maximum and local minimum should be at least 4 epochs. For clarification, the second condition implies that all the elements of the vector $range(w_{ij})$ should be less than a certain $D$, taking into consideration that the accuracy-related metric is measured in [0,100]. The first condition (i) guarantees a certain size of the window's interval to avoid noisy samples in terms of consecutive local maximums and minimums, while the second condition (ii) ensures the absence of big oscillations in the sampled values inside the window.

\subsection{The selection of the stop-window} 
\label{subsec:algo}

Following the above-mentioned mathematical foundations, we explain our proposal in Algorithm \ref{algorithm}. The algorithm is capable of selecting a stop-window of learning during the learning process, i.e. online, where the model is accurate enough.

First, in lines 1-9 of the algorithm we initialize the variables that we will use. Specifically, $min_{point}$ and $max_{point}$ denote the local minimum and local maximum respectively. Also, the variable \textit{epoch} denotes the current epoch learning process, while \textit{Swindow} and \textit{stopping} denote the initialization of the stop-window and the stopping state shows the Boolean condition according to which we will stop the training. The \textit{stopEpoch} variable denotes the selected epoch to stop the training process of the model. The variables $D$ and $N$ denote the maximum range of the window and minimum size of the window respectively. While a NN model is training after each epoch, we evaluate its performance on the test data using the $ImIoU$ metric and the value is returned to $f(epoch)$ variable. By calculating the $f'(epoch)$ and $f''(epoch)$, we check if the conditions to form local minimum (code lines 13-14) and local maximum (code lines 15-16) apply. In line 17, we check for the appearance of a local maximum prior to a local minimum of the performance metric and we define a window. Finally, in lines 19-22, we check the conditions to declare a window (w) as a stop-window ($Swindow$) and then we set the window to be qualified as a stop-window. Then, we keep the epoch where the model achieved its maximum $ImIoU$ value inside the $Swindow$, i.e. $stopEpoch$, otherwise we continue the training process of the NN. The function returns both the $Swindow$ and $stopEpoch$.

\begin{algorithm}[H]
 \label{algorithm}
\caption{Method of locating the Stop-window}

 $min_{point} = 0$\;
 $max_{point} = 0$\;
 maxEpochs = 200 \tcp{\scriptsize{\color{blue} can be changed to any value}}
 epoch = 0\;
 Swindow = []\;
 stopEpoch = 0 \;
 stopping = False\;
 $D = 2$ \tcp{\scriptsize{\color{blue}$D$ can be changed to any value in (0,2]}}
 $N = 4$ \tcp{\scriptsize{\color{blue}$N$ can be changed to any value in [2, max epoch - 1]}}
 \While{(!stopping) and (epoch $\leq$ maxEpochs)}{
   model@epoch = training() \tcp{\scriptsize{\color{blue} returns model @ current epoch}}
   $f(epoch)$ = testing(model@epoch) \tcp{\scriptsize{\color{blue} performance metric evaluation}} 
   \If{$f'(epoch) == 0$ and $f''(epoch) > 0$}{
   $min_{point} =$ epoch \tcp{\scriptsize{\color{blue}local minimum at this epoch}}
   }
   
   \If{$f'(epoch) == 0$ and $f''(epoch) < 0$}{ 
   $max_{point} =$ epoch \tcp{\scriptsize{\color{blue}local maximum at this epoch}} 
   }
   
   \If{$max_{point}$ < $min_{point}$}{ 
   w = [$max_{point}$, $min_{point}$] \tcp{\scriptsize{\color{blue}w denotes a window}} 
   \If{($size(w) \geq N$)) and ($\forall k$  $\in range(w): $ $ |k| < D$))}{
    Swindow = w\;
    stopEpoch = epoch where $max(f(epoch))\in$[Swindow]\;
    stopping = True\;
   }

   }

   epoch++\;
 }

 return Swindow, stopEpoch\;

\end{algorithm}

Note that, we stop the training once the first $Swindow$ is encountered. Besides, we return the \textit{stopEpoch}, i.e. the epoch of $Swindow$ in which the model achieved the best accuracy-related metric. Thus, the final model corresponds to the model trained until \textit{stopEpoch}. It is worth-mentioning that the policy rules to select a stop-window (lines 19-22) of the code can be reformulated.

\section{Evaluation}
\label{sec:evaluation}

This section describes our evaluation process and findings. Initially, we provide our evaluation protocol and then we show the utilized data and models.

\subsection{Evaluation Protocol}

We have established a standard protocol for training, validating, and testing each of the NN models we have selected. We used the Torch-Points3D \cite{Chaton2020Torch-Points3D:Clouds} framework to run the deep learning models, with Python 3.8.5, CUDA 10.2 and PyTorch 1.7.0 version. Also, our system configuration includes an 
Intel Core i9-10900 paired with 32GB RAM and a Quadro RTX 5000 with 16 GB, and the operating system is Ubuntu 20.04.

For the training, validating, and testing of the NN models, the split proposed by Chang et al. \cite{Chang2015ShapeNet:Repository} is used. Specifically, it comprises of 12137 point clouds for training, 1870 point clouds for validation, and 2874 point clouds for testing. Regarding the parameterization of the utilized NN, we set the batch size to be 16 and the optimizer to be Adam. We also use exponential learning rate decay and batch normalization on every epoch. Finally, following each epoch's training phase, all of the NN were validated and tested.

In the parameters of our early stopping algorithm we have set $D=2$, $N=4$, $maxEpochs=200$ and the monitored segmentation accuracy-related performance metric is $ImIoU$, as defined in \cite{Liu2019Relation-shapeAnalysis}.

\subsection{Data \& Models}

To evaluate our learning stopping strategy, we use the \textit{ShapeNet}\cite{Chang2015ShapeNet:Repository,Yi2016ACollections} part segmentation data, which is one of the most utilized datasets in the field. It comprises of 16881 3D point clouds categorised in 16 distinct classes of objects.

In addition, for the analysis of data we use four accurate deep learning models in the field of 3D point cloud segmentation: (i) \textbf{PointNet} \cite{Qi2017PointNet:Segmentation}, (ii) \textbf{PointNet++} \cite{Qi2017PointNet++:Space}, (iii) \textbf{KPConv} \cite{Thomas2019KPConv:Clouds} and (iv) \textbf{RSConv} \cite{Liu2019Relation-shapeAnalysis}. We utilize this selection of models because it consists of models with differences in their architecture and it encapsulates distinct approaches of part-segmentation, such as multi-layer perceptrons and convolutions.

\subsection{Analysis of our proposal}

Table \ref{tbl:stopwinsum} displays the proposed stop-windows accompanied with statistical measurements monitoring the $ImIoU$ metric for each model. The proposed stop-window in each model is denoted as \textbf{Swindow}. We display the average $ImIoU$ value of the stop-window, \textbf{SwAvg}, the standard deviation of it, \textbf{SwStd} and the maximum $ImIoU$ value, \textbf{SwMax}. We further show the maximum value of $ImIoU$ achieved in the whole learning process of 200 epochs, \textbf{Max}, the difference of \textbf{SwMax} with the \textbf{Max}, \textbf{SwMaxDiff} = $\frac{SwMax}{Max}$, i.e. 0 means highly different and 1 means no difference, and the difference of \textbf{SwAvg} versus the \textbf{Max}, \textbf{SwAvgDiff} = $\frac{SwAvg}{Max}$.

\begin{table}

\caption{Summary of the process of detecting stop-windows monitoring $ImIoU$.} 

\centering
\label{tbl:stopwinsum}
\resizebox{0.8\textwidth}{!}{%
\begin{tabular}{|c|c|c|c|c|c|}
\hline
{} & {\textbf{PointNet}} & {\textbf{PointNet++}} & {\textbf{KPConv}} & {\textbf{RSConv}} \\ \hline
\textbf{Swindow \tiny{(epochs)}}                       &     [38, 41]                         &    [22, 26]                             & [10, 14]                            &                    [12, 21]                     \\ \hline
\textbf{SwAvg \tiny{($ImIoU$)}}                   & 81.67                         & 83.48                           & 82.42                      & 84.42                      \\ \hline
\textbf{SwStd \tiny{($ImIoU$)}}                        & 0.10                          & 0.14                            & 0.24                       & 0.24                       \\ \hline
\textbf{SwMax \tiny{($ImIoU$)}}                        & 81.76                         & 83.63                           & 82.80                      & 84.82                      \\ \hline
\textbf{Max \tiny{($ImIoU$)}}                        & 84.24                         & 84.93                           & 84.22                      & 85.47                      \\ \hline\hline
\textbf{SwMaxDiff \tiny{($ImIoU$)}}             & 0.97                         & 0.98                           & 0.98                     & 0.99                     \\ \hline
\textbf{SwAvgDiff \tiny{($ImIoU$)}}             & 0.97                         & 0.98                           & 0.98                      & 0.99                      \\ \hline 

\end{tabular}
}

\end{table}

We can observe that the models stopped in the proposed stop-windows achieve approximately the same $ImIoU$ values as the best achieved in whole learning process of 200 epochs. For instance, it can be seen that the learning of KPConv model can be stopped at any epoch inside the window of $Swindow = [10, 14]$, which has a maximum value of $SwMax = 82.80$ with a deviation of only $SwStd = 0.24$. The difference between the window's average ($SwAvg = 82.42$) and max ($SwMax = 82.80$) from the general max recorded in 200 epochs ($Max = 84.22$) are $SwMaxDiff = 0.98$ and $SwAvgDiff = 0.98$ respectively, indicating that the stopping of training can be done early while having a highly accurate model (almost identical to the best $ImIoU$ obtained in 200 epochs). According to a recent performance benchmark shown in \cite{Zoumpekas2021BenchmarkingSegmentation}, KPConv needs a great amount of time to complete the learning process on ShapeNet dataset and specifically more time than its competitors.

\vspace{-0.2cm}
\hlbox{Observation 1}{\small{The process of learning becomes way less time consuming, while the test accuracy in all the analyzed models is approximately similar to the maximum accuracy achieved in 200 epochs. Thus, by employing our stopping algorithm the process can become much more time efficient, while the models still achieve high accuracy.}}
\vspace{-0.2cm}

\subsection{Comparison to conventional early stopping techniques}

We also compare our proposal with four common early stopping techniques, which deal with the overfitting issues of the models. The cross entropy segmentation loss in being monitored in the following conventional early stopping strategies: (i) \textbf{EarlyS1:} It stops the training process of the NN when the validation loss in the current epoch is higher than in the previous one; (ii) \textbf{EarlyS2:} It stops the training process when the validation loss in the current epoch is higher than the previous one by a 5\%; (iii) \textbf{EarlyS3:} A more advanced early stopping technique that considers a patience parameter. Patience refers to the number of epochs with no improvement in the monitored loss. For example a $patience=5$ indicates that the training will be stopped after 5 consecutive epochs of no improvement in the validation loss. In our case, we set $patience=2$; (iv) \textbf{EarlyS4:} It stops the training of the NN with $patience=3$.

\begin{table}[]
\caption{Comparison of our proposed technique versus four conventional early stopping techniques. We use dark grey and light grey cell colors to denote the best and the second best score per metric of each row respectively.}  \label{tbl:stopwindowVS}
\resizebox{\textwidth}{!}{%
\begin{tabular}{|c|c|c|c|c|c|c|}
\hline
{Model}                        & {Metric}        & \textbf{Our Technique} & \textbf{EarlyS1} & \textbf{EarlyS2} & \textbf{EarlyS3} & \textbf{EarlyS4} \\ \hline
\multirow{3}{*}{\textbf{PointNet}}   & {$ImIoU$}       & \cellcolor[gray]{0.65}81.76                           & 75.07                          & \cellcolor[gray]{0.85}79.54                          & 78.6                           & 77.26                          \\ \cline{2-7} 
                            & {$MaxDiff$}         & \cellcolor[gray]{0.65}0.97                            & 0.89                           & \cellcolor[gray]{0.85}0.94                           & 0.93                           & 0.92                           \\ \cline{2-7} 
                            & {$EffGain$} & Ep 39: 80.5(\%)                            & \cellcolor[gray]{0.65}Ep 3: 98.5(\%)                              & Ep 18: 91(\%)                                & \cellcolor[gray]{0.85}Ep 14: 93(\%)                                & Ep 15: 92.5(\%)                              \\ \hline
\multirow{3}{*}{\textbf{PointNet++}} & {$ImIoU$}       & \cellcolor[gray]{0.65}83.63                           & 78.66                          & 76.71                          & 80.94                          & \cellcolor[gray]{0.85}83.45                          \\ \cline{2-7} 
                            & {$MaxDiff$}         & \cellcolor[gray]{0.65}0.98                            & 0.93                           & 0.90                           & \cellcolor[gray]{0.85}0.95                           & \cellcolor[gray]{0.65}0.98                           \\ \cline{2-7} 
                            & {$EffGain$}     & Ep 26: 87(\%)                                 & \cellcolor[gray]{0.65}Ep 3: 98.5(\%)                              & \cellcolor[gray]{0.85}Ep 7: 96.5(\%)                              & Ep 11: 94.5(\%)                              & Ep 63: 86.5(\%)                              \\ \hline
\multirow{3}{*}{\textbf{KPConv}}     & {$ImIoU$}       & \cellcolor[gray]{0.85}82.80                           & 78.33                          & 81.04                          & 82.13                          & \cellcolor[gray]{0.65}83.15                          \\ \cline{2-7} 
                            & {$MaxDiff$}         & \cellcolor[gray]{0.85}0.98                            & 0.93                           & 0.96                           & \cellcolor[gray]{0.85}0.98                           & \cellcolor[gray]{0.65}0.99                           \\ \cline{2-7} 
                            & {$EffGain$}     & Ep 12: 94(\%)                                 & \cellcolor[gray]{0.65}Ep 3: 98.5(\%)                              & \cellcolor[gray]{0.85}Ep 7: 96.5(\%)                              & Ep 11: 94.5(\%)                              & Ep 63: 68.5(\%)                              \\ \hline
\multirow{3}{*}{\textbf{RSConv}}     & {$ImIoU$}       & \cellcolor[gray]{0.65}84.82                           & 81.84                          & \cellcolor[gray]{0.85}84.42                          & 81.87                          & \cellcolor[gray]{0.65}84.82                          \\ \cline{2-7} 
                            & {$MaxDiff$}         & \cellcolor[gray]{0.65}0.99                            & \cellcolor[gray]{0.85}0.96                           & \cellcolor[gray]{0.65}0.99                           & \cellcolor[gray]{0.85}0.96                           & \cellcolor[gray]{0.65}0.99                           \\ \cline{2-7} 
                            & {$EffGain$}     & Ep 20: 90(\%)                                 & \cellcolor[gray]{0.65}Ep 5: 97.5(\%)                              & Ep 21: 89.5(\%)                              & \cellcolor[gray]{0.85}Ep 6: 97(\%)                                & Ep 20: 90(\%)                                \\ \cline{1-7} 
\end{tabular}
}
\end{table}

Table \ref{tbl:stopwindowVS} shows a comparison of our proposed technique (\textbf{Our Technique}), which is the epoch that corresponds to the maximum $ImIoU \in$ Swindow, versus the four above-mentioned techniques. We denote $MaxDiff$ the division of the obtained $ImIoU$ in each strategy (stopEpoch, EarlyS1, EarlyS2, EarlyS3, EarlyS4) with the general max of $ImIoU$ of each model obtained in 200 epochs, and shows the difference of each metric versus the maximum obtained by the model in 200 epochs, i.e. 0 means highly different and 1 means no difference. Also, we denote $EffGain = (1 - \frac{Ep}{200}) * 100$, where $Ep$ is the epoch that we stopped the training according to each early stopping technique. $EffGain$ shows the efficiency gain of each model in each one of the early stopping strategies.

Observing the Table \ref{tbl:stopwindowVS}, we can note that in our proposed stop epoch (\textbf{SwMax}) the models achieved higher $ImIoU$ values than the models obtained from the other early stopping strategies. Regarding, the $MaxDiff$ metric, which is the division of $ImIoU$ obtained according to each strategy with the general max of $ImIoU$ obtained in 200 epochs, our strategy (\textbf{SwMax}) comes first in almost all the models, with the exception of KPConv ($MaxDiff = 0.98$) in which it comes second. For example, in RSConv and PointNet++ we achieve values equal to $MaxDiff = 0.99$ and $0.98$ respectively, indicating almost similar $ImIoU$ values with the maximum obtained in 200 epochs. Although in the $EffGain$ metric, our strategy comes last compared to its competitors, the obtained values of $EffGain$ are pretty close to the others, with the exception the PointNet ($EffGain = 80.5\%$).

\vspace{-0.2cm}
\hlbox{Observation 2}{\small{As the ultimate goal is to have a highly accurate model but also efficient in training time, in comparison to other techniques, our approach can be considered as the winner in selecting this model and an effective early stopping technique to be utilized in a point cloud segmentation task.}}
\vspace{-0.2cm}

\begin{figure}
    \centering
    \makebox[\linewidth][c]{%
    \includegraphics[width=1.2\textwidth]{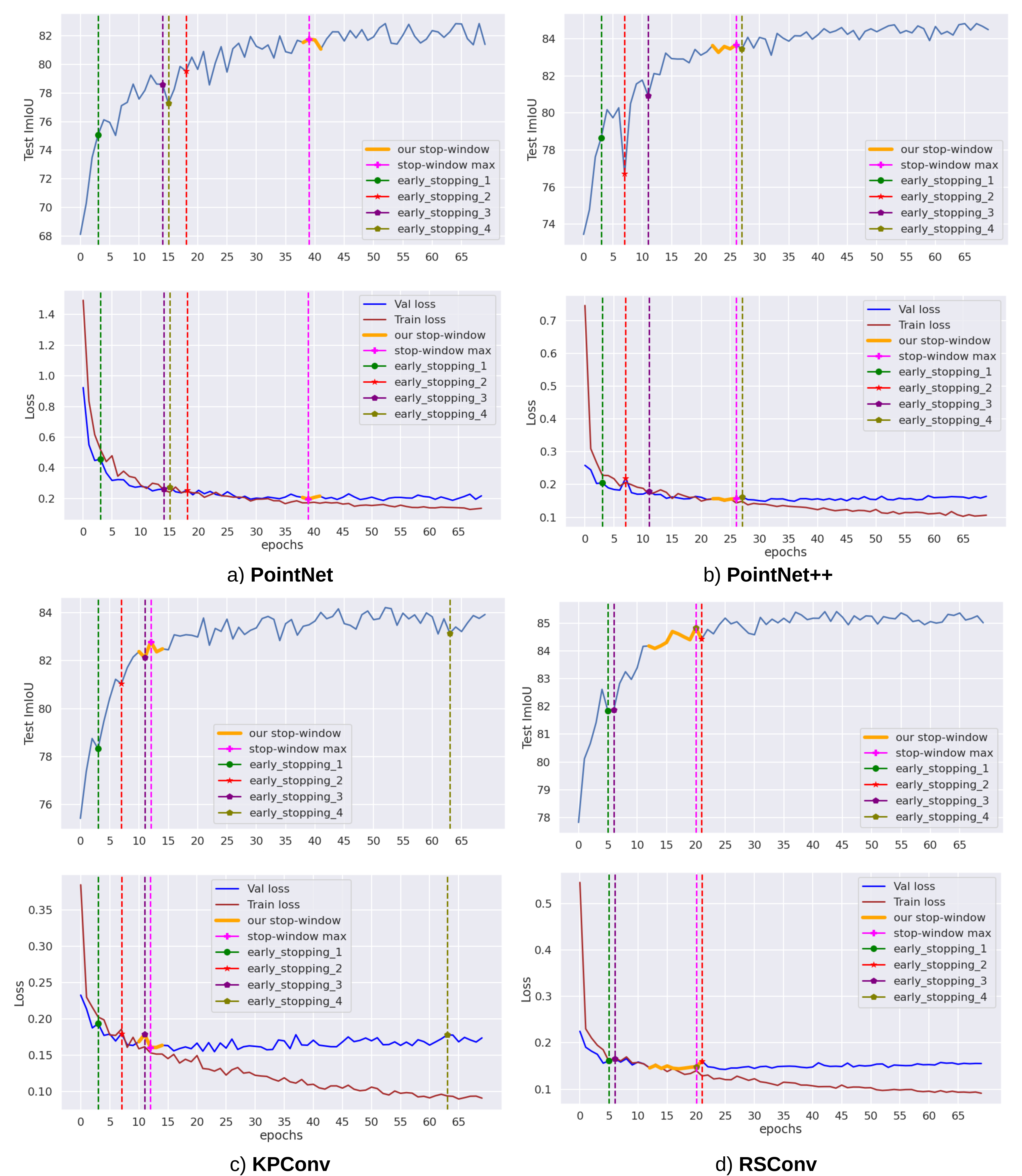}
    }
    \caption{Application of the conventional early stopping techniques versus our stop-window technique in four neural networks. } 
    \label{fig:results}
\end{figure}

Figure \ref{fig:results} shows a comparison of all the utilized early stopping strategies versus our proposal. In PointNet, we observe that our proposed stopping comes after the other early stopping techniques while achieving higher $ImIoU$ values. In the loss plot, we can see that our proposed stop-window takes place where the model starts to overfit, achieving lower loss values than its competitors. In PointNet++, we observe that our proposal behaves similar with the early stopping 4 strategy and they also detect better the spot where the overfitting of the model starts. Approximately the same behaviour appears in KPConv and RSConv models, with our proposed stop-windows competing well against their competitors.

\vspace{-0.2cm}
\hlbox{Observation 3}{\small{It seems that our proposal not only provides a higher segmentation accuracy ($ImIoU$) model than the other strategies but also a model which generalizes better in unseen data, i.e. the cross entropy loss is lower and the model learning is stopped right before it starts to overfit. In summary, our proposal is capable of returning a model highly accurate and efficient, which also competes well with the other strategies in identifying overfitting issues.}}
\vspace{-0.2cm}

\section{Conclusion}
\label{sec:conclusion}

This paper proposes an effective early stopping of point cloud NN based on mathematical foundations and focuses on the segmentation accuracy and efficiency rather than monitoring loss function values. Our results indicate a rather promising way of reducing the total time spent in the learning process of a NN, which can be easily utilized by a variety of researchers in the field. An individual can get highly accurate point cloud segmentation results in a time-efficient way. The comparison with several conventional early stopping techniques further justifies the effectiveness of our proposal. Our proposal is general enough to be utilized for monitoring any segmentation accuracy-related performance metric, either online, during the training of the network or after the training for data analysis of all the possible stop-windows. 

\section*{Acknowledgements}

\noindent\begin{minipage}{0.17\textwidth}
\includegraphics[width=\linewidth]{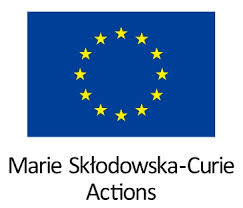}
\end{minipage}%
\hfill%
\begin{minipage}{0.83\textwidth}
This project has received funding from the European Union’s Horizon 2020 research and innovation programme under the Marie Skłodowska-Curie grant agreement No 860843.
\end{minipage}

\bibliography{samplepaper}   

\begin{thebibliography}{10}
\providecommand{\url}[1]{\texttt{#1}}
\providecommand{\urlprefix}{URL }
\providecommand{\doi}[1]{https://doi.org/#1}

\bibitem{Bai2021UnderstandingLabels}
Bai, Y., Yang, E., Han, B., Yang, Y., Li, J., Mao, Y., Niu, G., Liu, T.:
  {Understanding and Improving Early Stopping for Learning with Noisy Labels}
  (6 2021), \url{https://arxiv.org/abs/2106.15853v1}

\bibitem{Bello2020Review:Clouds}
Bello, S.A., Yu, S., Wang, C., Adam, J.M., Li, J.: {Review: Deep Learning on 3D
  Point Clouds}. Remote Sensing  \textbf{12}(11), ~1729 (5 2020).
  \doi{10.3390/rs12111729}

\bibitem{Caruana2000OverfittingStopping}
Caruana, R., Lawrence, S., Giles, L.: {Overfitting in Neural Nets:
  Backpropagation, Conjugate Gradient, and Early Stopping} (2000),
  \url{https://dl.acm.org/doi/10.5555/3008751.3008807}

\bibitem{Chang2015ShapeNet:Repository}
Chang, A.X., Funkhouser, T., Guibas, L., Hanrahan, P., Huang, Q., Li, Z.,
  Savarese, S., Savva, M., Song, S., Su, H., Xiao, J., Yi, L., Yu, F.:
  {ShapeNet: An Information-Rich 3D Model Repository}. arXiv  (12 2015),
  \url{http://arxiv.org/abs/1512.03012}

\bibitem{Chaton2020Torch-Points3D:Clouds}
Chaton, T., Chaulet, N., Horache, S., Landrieu, L.: {Torch-Points3D: A Modular
  Multi-Task Framework for Reproducible Deep Learning on 3D Point Clouds} (11
  2020). \doi{10.1109/3DV50981.2020.00029}

\bibitem{Garcia-Garcia2018ASegmentation}
Garcia-Garcia, A., Orts-Escolano, S., Oprea, S., Villena-Martinez, V.,
  Martinez-Gonzalez, P., Garcia-Rodriguez, J.: {A survey on deep learning
  techniques for image and video semantic segmentation}. Applied Soft Computing
   \textbf{70},  41--65 (9 2018). \doi{10.1016/j.asoc.2018.05.018}

\bibitem{Guo2020DeepSurvey}
Guo, Y., Wang, H., Hu, Q., Liu, H., Liu, L., Bennamoun, M.: {Deep Learning for
  3D Point Clouds: A Survey}. IEEE Trans. on Pattern Analysis and Machine
  Intelligence  \textbf{43}(12),  4338--4364 (12 2021).
  \doi{10.1109/TPAMI.2020.3005434}

\bibitem{Hegde2021PIG-Net:Segmentation}
Hegde, S., Gangisetty, S.: {PIG-Net: Inception based deep learning architecture
  for 3D point cloud segmentation}. Computers and Graphics (Pergamon)
  \textbf{95},  13--22 (2021). \doi{10.1016/j.cag.2021.01.004}

\bibitem{Liu2019DeepSurvey}
Liu, W., Sun, J., Li, W., Hu, T., Wang, P.: {Deep learning on point clouds and
  its application: A survey} (2019). \doi{10.3390/s19194188}

\bibitem{Liu2019Relation-shapeAnalysis}
Liu, Y., Fan, B., Xiang, S., Pan, C.: {Relation-shape convolutional neural
  network for point cloud analysis} (4 2019). \doi{10.1109/CVPR.2019.00910}

\bibitem{Prechelt2012EarlyWhen}
Prechelt, L.: {Early Stopping — But When?} Lecture Notes in Computer Science
  (including subseries Lecture Notes in Artificial Intelligence and Lecture
  Notes in Bioinformatics)  \textbf{7700 LECTURE NO},  53--67 (2012).
  \doi{10.1007/978-3-642-35289-8{\_}5}

\bibitem{Qi2017PointNet:Segmentation}
Qi, C.R., Su, H., Mo, K., Guibas, L.J.: {PointNet: Deep learning on point sets
  for 3D classification and segmentation} (2017). \doi{10.1109/CVPR.2017.16}

\bibitem{Qi2017PointNet++:Space}
Qi, C.R., Yi, L., Su, H., Guibas, L.J., Li, C.R.Q., Hao, Y., Leonidas, S.,
  Guibas, J.: {PointNet++: Deep hierarchical feature learning on point sets in
  a metric space} (2017). \doi{10.5555/3295222}

\bibitem{Thomas2019KPConv:Clouds}
Thomas, H., Qi, C.R., Deschaud, J.E., Marcotegui, B., Goulette, F., Guibas, L.:
  {KPConv: Flexible and deformable convolution for point clouds} (4 2019).
  \doi{10.1109/ICCV.2019.00651}

\bibitem{Yan2020PointASNL:Sampling}
Yan, X., Zheng, C., Li, Z., Wang, S., Cui, S.: {PointASNL: Robust Point Clouds
  Processing Using Nonlocal Neural Networks With Adaptive Sampling} (2020).
  \doi{10.1109/cvpr42600.2020.00563}

\bibitem{Yi2016ACollections}
Yi, L., Kim, V.G., Ceylan, D., Shen, I.C., Yan, M., Su, H., Lu, C., Huang, Q.,
  Sheffer, A., Guibas, L.: {A scalable active framework for region annotation
  in 3D shape collections}. ACM Trans. on Graphics  \textbf{35}(6) (2016).
  \doi{10.1145/2980179.2980238}

\bibitem{Ying2019AnSolutions}
Ying, X.: {An Overview of Overfitting and its Solutions}. Journal of Physics:
  Conf. Series  \textbf{1168}(2),  022022 (2 2019).
  \doi{10.1088/1742-6596/1168/2/022022}

\bibitem{Zhang2019ACloud}
Zhang, J., Zhao, X., Chen, Z., Lu, Z.: {A Review of Deep Learning-Based
  Semantic Segmentation for Point Cloud}. IEEE Access  \textbf{7},
  179118--179133 (2019). \doi{10.1109/ACCESS.2019.2958671}

\bibitem{Zoumpekas2022CLOSED:Learning}
Zoumpekas, T., Molina, G., Puig, A., Salam{\'{o}}, M.: {CLOSED: A Dashboard for
  3D Point Cloud Segmentation Analysis using Deep Learning}. Proc. of the 17th
  Int. Joint Conf. on Computer Vision, Imaging and Computer Graphics Theory and
  Applications pp. 403--410 (2022). \doi{10.5220/0010826000003124}

\bibitem{Zoumpekas2021BenchmarkingSegmentation}
Zoumpekas, T., Molina, G., Salam{\'{o}}, M., Puig, A.: {Benchmarking Deep
  Learning Models on Point Cloud Segmentation}. In: Artificial Intelligence
  Research and Development, vol.~339, pp. 335--344 (10 2021).
  \doi{10.3233/FAIA210152}

\end{thebibliography}

\end{document}